\documentclass[epj]{svjour}
\usepackage[dvips]{graphicx}

\begin{document}

\markboth{Juan P. Herrera and Pedro A. Pury}
{Statistical keyword detection in literary corpora}

\title{Statistical Keyword Detection in Literary Corpora}

\author{
Juan P. Herrera\thanks
{Present address:
Argentina Software Development Center (ASDC),
Intel Software, C\'ordoba, Argentina
(e-mail: juan.herrera@intel.com).}
\and
Pedro A. Pury\thanks
{Corresponding author
(e-mail: pury@famaf.unc.edu.ar).}
}

\institute{
Facultad de Matem\'atica, Astronom\'\i a y F\'\i sica,
Universidad Nacional de C\'ordoba, \\
Ciudad Universitaria, X5000HUA C\'ordoba, Argentina}

\date{Received: 1st May 2007 /
Received in final form 15 February 2008 \\
$\copyright$ EDP Sciences, Societ{\`a} Italiana di Fisica,
Springer-Verlag 2008}


\abstract{
Understanding the complexity of human language requires
an appropriate analysis of the statistical distribution
of words in texts. We consider the information retrieval problem
of detecting and ranking the relevant words of a text by means
of statistical information referring to the {\em spatial} use
of the words. Shannon's entropy of information is used as
a tool for automatic keyword extraction.
By using {\em The Origin of Species} by Charles Darwin as
a representative text sample, we show the performance of
our detector and compare it with another proposals in
the literature. The random shuffled text receives special
attention as a tool for calibrating the ranking indices.
}


\maketitle

\section{Introduction}
\label{sec:intro}

Data mining for texts is a well-established area of natural language
processing~\cite{MS99}. Text mining is the computerised extraction
of useful answers from a mass of textual information by machine
methods, computer-assisted human ones, or a combination of both.
A key problem in text mining is the extraction of keywords
from texts for which no {\em a priori} information is available.
The problem of unsupervised extraction of relevant words from their
statistical properties was first addressed by Luhn~\cite{Luh58},
who based his method on Zipf's analysis of frequencies~\cite{Zipf}.
This analysis consists of counting the number of occurrences
of each distinct word in a given text, and then generating a list
of all these words ordered by decreasing frequency. In this list,
each word is identified by its position or {\em Zipf's rank}
in the list. The empirical observation of Zipf was that the
frequency of occurrence of the $r$--th rank in the list is
proportional to $r^{-1}$ ({\em Zipf's law}). Luhn proposed
the crude approach of excluding the words at both ends of
the Zipf's list and considering as keywords the remaining cases.
The limitations of Luhn's approach are known in the
literature~\cite{SM83}.

The main goal of this work is to investigate unsupervised
statistical methods for detecting keywords in literacy texts
beyond the simple counting of word occurrences.
In order to obtain statistically significant results we restrict
our work to a large book, which can be used as a corpus what
is thematically consistent throughout its entire length.
We are searching for relevance according to the text's context, but
we will only use statistical information about the {\em spatial}
use of the words in a text.

Particularly, the measure of content of information for each word
can be made by Shannon's entropy.
In the physics literature, we can find several applications of
the entropy concept to linguistics and natural language like
{DNA} sequences analysis~\cite{MBG+95,SBG+99,GBC+02},
long-range correlations measurements~\cite{EP94,EPA95},
language acquisition~\cite{CCGG99},
authorship disputes~\cite{CMH97,YPYG03},
communication modelling~\cite{Can05},
and statistical analysis of the linguistic role
of words in corpora~\cite{MZ02}.

The organisation of the remainder of the article is as follows.
In Sec.~\ref{sec:Darwin} we first introduce the corpus used
as a representative sample throughout this work.
Later, in Sec.~\ref{sec:cluster} we review the algorithms
proposed in the literature based on the analysis
of the statistical distribution of words in a text.
Then, in Sec.~\ref{sec:random} we discuss the behaviour
of the indices in random texts.
By using Shannon's entropy, in Sec.~\ref{sec:entropy}
we propose another index based on the information content
of the sequence of occurrences of each word in the text.
In Sec.~\ref{sec:benchmark} we use the glossary of the corpus
for measuring the performace of each index as keyword
detector. Finally, in Sec.~\ref{sec:fin} we present a summary
of the work. Besides, mathematical details are given in appendices.
In Appendix~\ref{app:pascal} we review the geometrical distribution,
useful to random texts, and in Appendix~\ref{app:entropy}
we calculate the entropy of a random text.

\section{Representative Corpus Sample}
\label{sec:Darwin}

For our study, we will use a prototypical real text, {\em i.e.},
{\em ``On the Origin of Species by Means of Natural Selection,
or The Preservation of Favoured Races in the Struggle
for Life''}~\cite{Gut} (usually abbreviated to
{\em The Origin of Species}) by Charles Darwin (1859).
The book was written with the vocabulary of a nineteenth-century
naturalist but with fluid prose, combining first--person
narrative with scholarly analysis.

For the preparation of our working corpus we first withdrew
any punctuation symbol from the text, mapped all words to
uppercase and then used the simple tokenization method based
on whitespaces~\cite{MS99}.
We draw a distinction between a word token versus a word type.
For our convenience, we define a word type as any different string
of letters between two whitespaces. Thus, for our elementary
analysis, words like {\small INSTINCT} and {\small INSTINCTS}
correspond to different word types in our corpus. On the other hand,
a word token is each individual occurrence of a given word type.
When the context refers to a particular word type, we will use
indistinctly ``word token'' or simply ``token'' to refer
to an individual occurrence of the word type in the text.

The relevant words have not been explicitly defined in
Darwin's book, with exception of a glossary appended at the
end of the work. Therefore, the table of contens in the beginning,
the glossary and the analytical index, also inserted at the end,
were removed from our corpus. By doing this, we avoid introducing
obvious bias for the words used in these parts.
Thus, the prepared corpus includes $94 \%$ of material from the
original Darwin's book and has $192,665$ word tokens and $8,294$
word types. In addition, the corpus contains $842$ paragraphs
distributed in $16$ chapters.

The glossary of the principal scientific terms used in the book,
prepared by Mr. W.S. Dallas, and the analytical index, both appended
at the end of the book, were written using $2,418$ word types.
If we do not consider the function words, still remain $1,679$ word
types ($20 \%$ of the book's lexicon).
With this information,, we prepared by hand a customed version
of the glossary, by selecting $283$ word types ($3.4 \%$ of the
lexicon) with frequencies of occurrence greater than 9.
We have avoided word types with less than 9 occurrences because
we cannot extract any significant statistics from data
obtained using such small sets.
Thus, the criterion for selection was rather more arbitrary,
but we think that all selected words are pertinent to
the book's context.
Our prepared version of the glossary will be used later
to evaluate the retrieval capabilities of different
keyword extractors.

\section{Clustering as criterion for relevance of words}
\label{sec:cluster}

\begin{figure}
\begin{center}
\includegraphics[clip,width=0.45\textwidth]{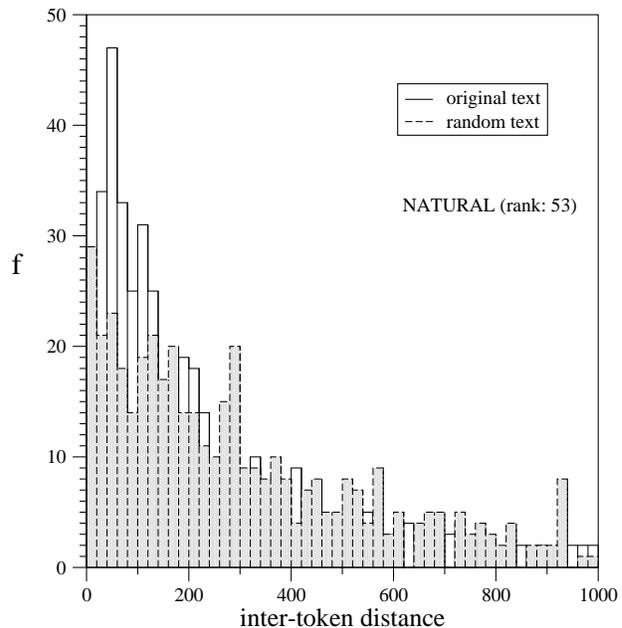}
\end{center}
\caption
{\small Histogram of frequencies of distances between occurrences
of {\small NATURAL} (Zipf's rank 53) in Darwin's corpus.
}
\label{fig:histogram}
\end{figure}

\begin{figure}
\begin{center}
\includegraphics[clip,width=0.45\textwidth]{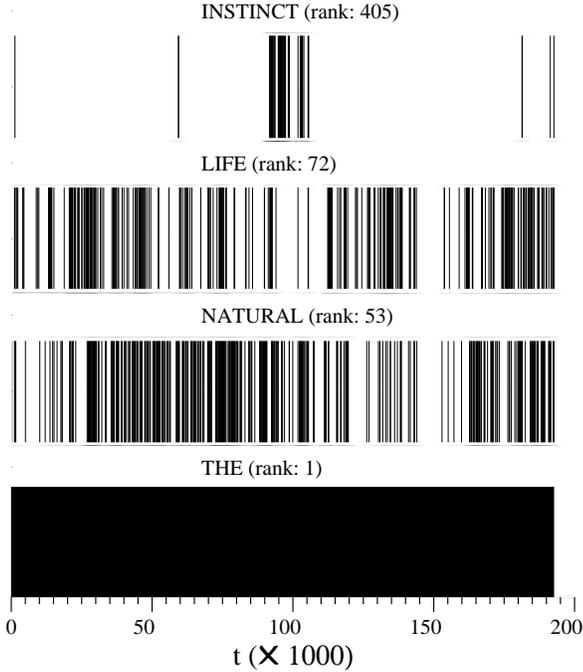}
\end{center}
\caption
{\small Absolute positions ($t$) in the text, counted from the
beginning of Darwin's corpus, of the word types:
{\small THE}   ($13,414$ occurrences),
{\small NATURAL} ($475$ occurrences),
{\small LIFE}    ($326$ occurrences), and
{\small INSTINCT} ($64$ occurrences).
To draw the picture, we set a very thin vertical line
(of arbitrary height) at the position of each occurrence.
}
\label{fig:barcodes}
\end{figure}

The attraction between words is a phenomenon that plays an important
role in both language processing and acquisition, and it has been
modeled for information retrieval and speech recognition
purposes~\cite{BBL97,NW97}.
Empirical data reveals that the attraction between words decays
exponentially, while stylistic and syntactic constraints create
a repulsion between words that discourages close occurrences.
In Fig.~(\ref{fig:histogram}) we have plotted the histogram
of absolute frequencies of distances between nearest neighbour
tokens of the word type {\small NATURAL} in Darwin's corpus.
For long distances, Fig.~(\ref{fig:histogram}) qualitatively suggests
an exponential tail, but for very short distances the frequencies
decay abruptly.
Also in Fig.~(\ref{fig:histogram}) we have superimposed the
histogram of a random shuffled version of the corpus where
we can qualitatively see an exponential decay for all distances.
The attraction--repulsion phenomenon is more emphasized for
relevant words than for common words, which have less
syntactic penalties for close co-occurrence.
Therefore, the spatial distributions of relevant words
in the text are inhomogeneous and these words gather
together in some portions of the text forming clusters.
The clustering phenomenon can be visualised in
Fig.~\ref{fig:barcodes} where we have plotted the absolute
positions of four different word types from Darwin's corpus
in a ``bar code'' arrangement.
The clustering becomes manifest in the patterns of {\small NATURAL},
{\small LIFE}, and {\small INSTINCT} in spite of their different
numbers of occurrences. In contrast, {\small THE} (the more frequent
word in the English language) has no apparent clustering.

Recently, the assumption that highly relevant words should be
concentrated in some portions of the text was used for
searching relevant words in a given text.
In the following two subsections, we briefly review the indices
of relevance of words proposed by Ortu\~no et al.~\cite{OCB+02}
and Zhou and Slater~\cite{ZS03}, which are based on the spatial
distribution of words in the text.

\subsection{$\sigma$--index}
\label{subsec:sigma}

To study the spatial distribution of a given word type in a text,
we can map the occurrences of the corresponding word tokens into
a time series. For this task, we denote by $t_i$ the absolute
position in the corpus of the $i$--th occurrence of a word token.
Thus, we obtain the sequence
$\{ t_0, t_1, \dots, t_n, t_{n+1} \}$,
where we are assuming that there are $n$ word tokens.
We have additionally included the boundaries of the corpus,
defining $t_0 = 0$ and $t_{n+1} = N+1$, where $N$ is the total
number of tokens in the corpus, in order to take into account
the space before the first occurrence of a word token and
the space after the last occurrence of a token~\cite{ZS03}.

Given the sequence of inter--token distances
$$\{ t_1-t_0, t_2-t_1, \dots, t_n-t_{n-1}, t_{n+1}-t_n \} \,,$$
the average distance between two successive word tokens is given by
\begin{equation}
\mu = \frac{1}{n+1} \,
      \sum_{i=0}^{n} \left( t_{i+1} - t_{i} \right)
    = \frac{N+1}{n+1} \;,
\label{eq:mu}
\end{equation}
and the sample standard deviation of the set of spacings between
nearest neighbour word tokens ($t_{i+1}-t_{i}$) is by definition
\begin{equation}
s = \sqrt{\frac{1}{n-1} \,
    \sum_{i=0}^{n} \left( (t_{i+1}-t_{i})-\mu \right)^2} \;.
\label{eq:s}
\end{equation}

To eliminate the dependence on the frequency of occurrence
for different word types, in Ref.~\cite{OCB+02} the authors suggest
to normalise the token spacings, {\em i.e.}, to measure them in
units of their corresponding mean value. Thus, we define
\begin{equation}
\sigma = \frac{s}{\mu} \;.
\label{eq:sigma}
\end{equation}

Given that the standard deviation grows rapidly when the
inhomogeneity of the distribution of spacing $t_{i+1}-t_{i}$
increases, Ortu\~no et al.~\cite{OCB+02} proposed $\sigma$ as
an indicator of the relevance of the words in the analysed text.
In many cases, empirical evidence vindicates that large $\sigma$
values generally correspond to terms relevant to the text considered,
and that common words have associated low values of $\sigma$.
However, Zhou and Slater~\cite{ZS03} pointed out that $\sigma$-index
has some weaknesses.
First, several obviously common (relevant) words have relative
high (low) $\sigma$ values in several texts.
Second, the index is not stable in the sense that it can be
strongly affected by the change of a single occurrence position.
Third, high values of $\sigma$ do not always imply a cluster
concentration. A big cluster of words can be splitted into smaller
clusters without substantial change in the $\sigma$ value.

\subsection{$\Gamma$--index}
\label{subsec:gamma}

The $\sigma$-index is only based on the spacing between
nearest-neigh\-bour word tokens. To improve the performance
in the searching for relevance, Zhou and Slater~\cite{ZS03}
introduced a new index that uses more information from the
sequence of occurrences $\{ t_0, t_1, \dots, t_n, t_{n+1} \}$.
For this task, these authors consider the spacings
$w_i = t_{i}-t_{i-1}$, with $i = 1, \dots, n+1$, and define
the {\em average separation} around the occurrence at $t_i$ as
\begin{equation}
d(t_{i}) = \frac{w_{i+1} + w_{i}}{2}
         = \frac{t_{i+1}-t_{i-1}}{2}, \qquad i=1,\ldots,n \,.
\label{eq:separa}
\end{equation}
The position $t_i$ is said to be a cluster point if $d(t_i) < \mu$.
The new suggestion is that the relevance of a word in a given
text is related to the number of cluster points found in it.
Thus, in order to measure the degree of clusterization, the local
{\em cluster index} at position $t_i$ is defined by
\begin{equation}
\gamma(t_{i}) =
\left\{
\begin{array}{lcl}
\displaystyle\frac{\mu - d(t_{i})}{\mu} &&
\textrm{ if $t_{i}$ is a cluster point}\\
0 &&
\textrm{ otherwise}
\end{array}
\right. \;.
\label{eq:cluster}
\end{equation}
Finally, a new index to measure relevance is obtained
from the average of all {\em cluster indices} corresponding
to a given word type
\begin{equation}
\Gamma = \frac{1}{n}\sum_{i=1}^n \gamma(t_{i}) \;.
\label{eq:Gamma}
\end{equation}
$\Gamma$-index is more stable than $\sigma$, but it is
still based on local information and is computationally
more time consuming to evaluate than $\sigma$.

\section{Random text and shuffled words}
\label{sec:random}

In a completely random text we have an uncorrelated
sequence of tokens, and a word type $w$ is only characterised
by its relative frequency of occurrence ($p_w$).
Thus, a random text can be generated by picking successively
tokens by chance in such a way that at each position
the probability of finding a token, corresponding to the
word type $w$, is $p_w$. Obviously, $\sum_w p_w = 1$.
For the word type $w$, we have in this manner defined a binomial
experiment where the probability of success (occurrence)
at each site in the text is $p_w$, and the probability
of failure (non-occurrence) is $(1-p_w)$.
Therefore, the distribution of distances between nearest neighbour
tokens corresponding to the same word type is geometrical.
In Appendix~\ref{app:pascal}, we have compiled some results
of the geometrical distribution that are useful for our next
analyses.

Besides, its worth as comparative standard, the theoretical
random text has the virtue of being analytically tractable.
Also, from an empirical point of view, there is a workable
fashion for building a random version of a corpus.
In an actual corpus the probabilities of occurrence $p$
are estimated from the relative frequencies $n/N$,
where $n$ is the number of tokens  corresponding to a given
word type and $N$ is the total number of tokens in the corpus.
A random version of the text can be obtained by shuffling
or permuting all the tokens.
The random shuffling of all the words has the effect of rescasting
the corpus into a nonsensical realization, keeping the same original
tokens without discernible order at any level.
However, both the Zipf's list of ranks and the frequency of
occurrence of each word type are kept intact.

The important point that we want to stress here is that the indices
of relevance defined in the previous section are functions of
the frequencies of occurrence of each word type.
Thus, in a random text the values of these indices change with
$p$, which has nonsense. In a truly random text, there are not
relevant words. Therefore, to eliminate completely the dependence
on frequency we need to renormalise the indices with their values
in the random version of the corpus.

\subsection{Renormalised $\sigma$--index}
\label{subsec:hsigma}

For a given probability distribution, $\sigma$ is defined from the
second-- ($\mu_2$) and first--order ($\mu_1$) cumulant by
$\sqrt{\mu_2}/\mu_1$. Thus, from Eq.~(\ref{cumulants})
in Appendix~\ref{app:pascal} we find that in a random text
the value of $\sigma$--index is given by
\begin{equation}
\sigma_{\mbox{ran}} = \sqrt{1-p} \;.
\label{eq:sigmaran}
\end{equation}
Hence, we renormalise the index to eliminate this dependence on
relative frequency defining
\begin{equation}
\sigma_{\mbox{nor}} = \frac{s}{\mu} \frac{1}{\sqrt{1-p}}\;.
\label{eq:sigmanor}
\end{equation}
For texts as large as corpora the importance of normalisation
factor given by Eq.~(\ref{eq:sigmaran}) becomes negligible.
For example, in Darwin's corpus, $N = 192,665$, and for the
most frequent word type ({\small THE}) we have $n = 13,414$
($n/N = 0.0696$).
Thus, in the less significant case (the lowest value for
$\sigma_{\mbox{ran}}$) $\sigma_{\mbox{ran}} = 0.965$,
whereas $\sigma_{\mbox{ran}} = 1$ for $p=0$.
However, for shorter texts the significance of the normalisation
may become critical and the values of $\sigma$ and
$\sigma_{\mbox{nor}}$ may be very different for any word type.

\begin{figure}
\begin{center}
\includegraphics[clip,width=0.45\textwidth]{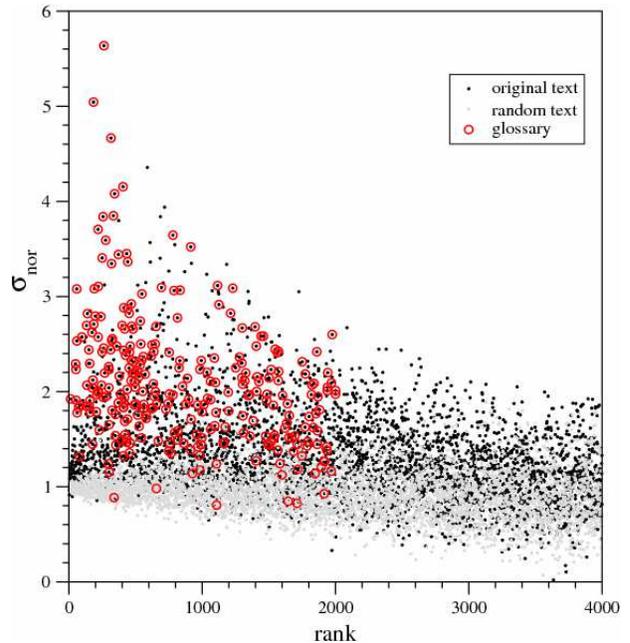}
\end{center}
\caption
{\small Renormalised $\sigma$--index vs.\ Zipf's rank for each
word in Darwin's corpus (the first 4000 ranks).
We have also plotted superposed the random version of the text (grey)
and we have marked by open circles the words corresponding
to our prepared glossary (red online).}
\label{fig:sigma}
\end{figure}

In Fig.~\ref{fig:sigma} we plot the values of $\sigma_{\mbox{nor}}$
for the first $4000$ ranks in the Zipf's list of Darwin's corpus.
The random version of the corpus is also plotted in the same
graph. The ``cloud of points'' corresponding to the random text
is distributed around the unitary value of $\sigma_{\mbox{nor}}$,
but the width of the ``cloud'' grows with rank. This behaviour is
due to the fact that the frequency of occurrence decreases as the
rank increases (Zipf's law), therefore the statistics get worse.
The words of our preparated version of the glossary are marked
by open circles in Fig.~\ref{fig:sigma}.
From Fig.~\ref{fig:sigma}, it is appreciable that most of
the glossary words have high values of $\sigma_{\mbox{nor}}$.
%

\subsection{Renormalised skewness}
\label{subsec:skew}

As in the case of $\sigma$, any cumulant contains partial
information of the spatial distribution of words. Skewness
is a parameter that describes the asymmetry of a distribution.
Mathematically, the skewness is measured using the second-- ($\mu_2$)
and third--order ($\mu_3$) cumulant of the distribution according
to $\kappa = \mu_3 / \mu_2^{3/2}$.
Given that the distances between nearest neighbour tokens are
positive defined, the corresponding distribution has positive skew,
{\em i.e.}, the upper tail is longer than the the lower tail
(see Fig.~\ref{fig:histogram}).

From Eq.~(\ref{cumulants}), we find that in a random text the
skewness of the distribution of distances between nearest neighbour
tokens is given by
\begin{equation}
\kappa_{\mbox{ran}} = \displaystyle\frac{2-p}{\sqrt{1-p}} \;;
\label{eq:kapparan}
\end{equation}
Thus, the skewness also depends on the relative frequency of
occurrence, $p$, in the random case. However, this dependence
is also negligible for a corpora. In Darwin's corpus we obtain
$\kappa_{\mbox{ran}} = 2.001$ for the largest value $p = 0.0696$
(the relative frequency of the word type {\small THE}),
whereas $\kappa_{\mbox{ran}} = 2$ for $p=0$.

\begin{figure}
\begin{center}
\includegraphics[clip,width=0.45\textwidth]{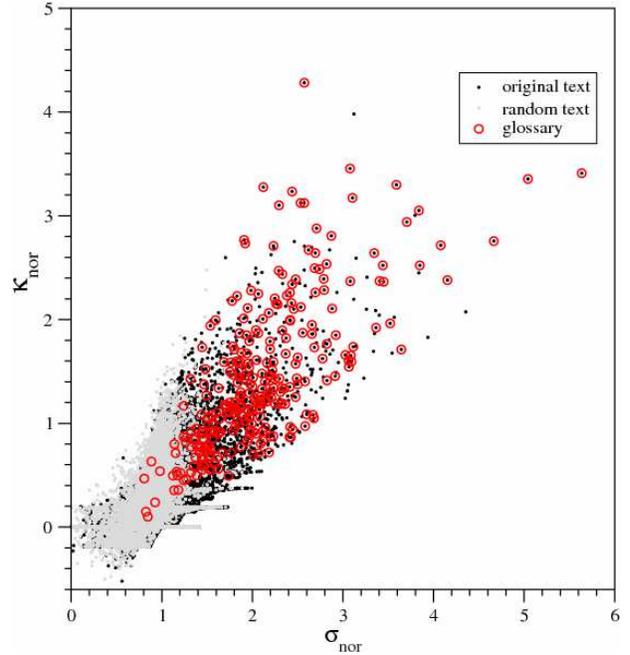}
\end{center}
\caption
{\small Renormalised $\kappa$--index vs.\ $\sigma$--index
for {\em all} words in Darwin's corpus.
We have also plotted superposed the random version of the text
(grey) and we have marked by open circles the words corresponding
to our prepared glossary (red online).}
\label{fig:sigka}
\end{figure}

As a consequence, we can define another renormalised quantity
as we did with the $\sigma$--index. Thus, to eliminate the
dependence on the relative frequency of occurrence in the
random case, we write
\begin{equation}
\kappa_{\mbox{nor}} =
\frac{\mu_3}{\mu_2^{3/2}} \frac{\sqrt{1-p}}{2-p}\;.
\label{eq:kappareno}
\end{equation}
$\kappa_{\mbox{nor}}$ can also be used for measuring relevance.
However, the finite-size effects of the texts are more pronounced
for higher order cumulants.
We now use both cumulants $\sigma_{\mbox{nor}}$ and
$\kappa_{\mbox{nor}}$ to construct a bi-dimensional graph
for the corpus. In this manner, in Fig.~\ref{fig:sigka}
we plot the the pairs $(\sigma_{\mbox{nor}},\kappa_{\mbox{nor}})$
for {\em all} words in Darwin's corpus. In this graph, the
``cloud of points'' corresponding to the random text is distributed
around the pair of values $(1,1)$, while the region defined by
$\sigma_{\mbox{nor}} > 2$ and $\kappa_{\mbox{nor}} > 2$ has
almost none.
The upper right corner of the graph concentrates almost all
the points corresponding to the glossary.
Figure~\ref{fig:sigka} gives us immediate insight into the
distribution of distances between nearest neighbour tokens,
and provides us a powerful tool for determining keywords.

\subsection{Renormalised $\Gamma$--index}
\label{subsec:hgamma}

As we did with the $\sigma$--index, we need to calculate $\Gamma$
for a word type which appears in a random text with relative
frequency $p$.
For this task, we calculate the average of the random variable
$\gamma$ defined in Eq.~(\ref{eq:cluster}) in a random text.
From Eq.~(\ref{gammaran}) in Appendix~\ref{app:pascal} we obtain
\begin{equation}
\Gamma_{\mbox{ran}} = \displaystyle\frac{1}{2} \,h \,(h-1)
\,(1-p)^h \,((1-p) + (1-p)^{-1} - 2) \;,
\label{eq:Gammaran}
\end{equation}
where $h = \mbox{Int}[2/p]$.
In this case, the dependence on $p$ is even more complicated than
previous cases. This observation is absent from Ref.~\cite{ZS03}.
Zhou and Slater only calculate the value of $\Gamma$ for the Poisson
distribution: $\Gamma = 2 \,e^{-2}$ (see Eq.~(\ref{Poisson})
in Appendix~\ref{app:pascal}), which is constant ($\approx 0.271$).
Also in this case, the dependence on $p$ is negligible for corpora.
In Darwin's corpus we obtain $\Gamma_{\mbox{ran}} = 0.261$ for
the largest value of $p = 0.0696$ (the relative frequency of
the word type {\small THE}), whereas $\Gamma_{\mbox{ran}}
\approx 0.271$ in the limit $p \rightarrow 0$
(see Appendix~\ref{app:pascal}).

\begin{figure}
\begin{center}
\includegraphics[clip,width=0.45\textwidth]{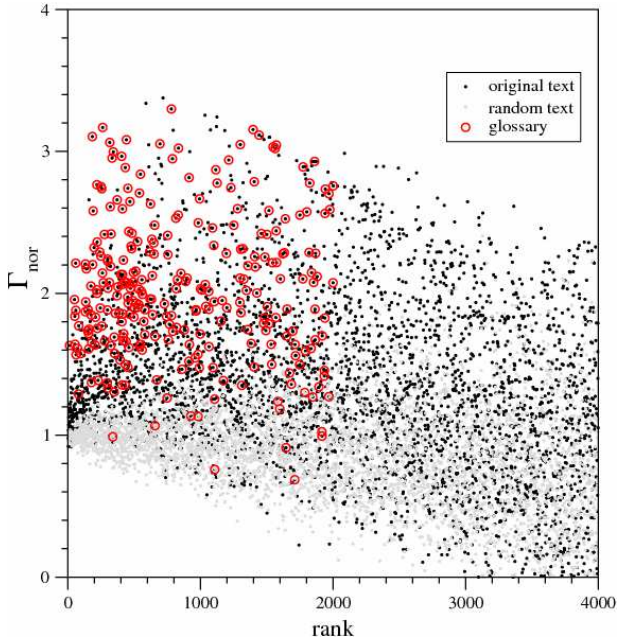}
\end{center}
\caption
{\small Renormalised $\Gamma$--index vs.\ Zipf's rank for each
word in Darwin's corpus (the first 4000 ranks).
We have also plotted superposed the random version of the text (grey)
and we have marked by open circles the words corresponding to our
prepared glossary (red online).}
\label{fig:gamma}
\end{figure}

Now, as in the other cases, we define from Eqs.~(\ref{eq:Gamma})
and~(\ref{eq:Gammaran}) a renormalised index by
$\Gamma_{\mbox{nor}} = \Gamma /\Gamma_{\mbox{ran}}$.
In Fig.~\ref{fig:gamma} we plot the values of $\Gamma_{\mbox{nor}}$
for the first $4000$ ranks in the Zipf's list of Darwin's corpus.
The ``cloud of points'' corresponding to the random text
is distributed around the unitary value, but the width of
the ``cloud'' grows with rank faster than in the case of
$\sigma_{\mbox{nor}}$.
The words corresponding to the glossary have systematically
high values of $\Gamma_{\mbox{nor}}$.
%

\section{Entropy of token distributions}
\label{sec:entropy}

Claude Shannon introduced the concept of {\em entropy of information}
in 1948~\cite{SW49}. Mapping a discrete information source on a set
of possible events whose probabilities of occurrences
are $p_1, p_2, \dots, p_P$, Shannon constructed a measure of
information and uncertainty, $S(p_1, p_2, \dots, p_P)$,
requiring the following properties:
\begin{enumerate}

\item $S$ should be continuous in the $\{p_i\}$.

\item For the iso-probability case, $p_i = 1/P$, $S$ should be
a monotonic increasing function of $P$.

\item If the set $p_1, p_2, \dots, p_P$ is broken down
into two subsets with probabilities
$w_1 = p_1 + \dots + p_k$ and $w_2 = p_{k+1} + \dots + p_P$,
then we must have the following composition law
$S(p_1, \dots p_N) = S(w_1,w_2) +
w_1 \,S(p_1/w_1, \dots p_k/w_1) \\+
w_2 \,S(p_{k+1}/w_2, \dots p_{P}/w_2)$.
\end{enumerate}
The only $S$ satisfying the three above assumptions
is of the form
\begin{equation}
S(p_1, p_2, \dots, p_P) = -K \sum_{i=1}^P p_i \,\log p_i \,,
\label{eq:Shannon}
\end{equation}
where $K$ is a positive constant.

A literary corpus can be divided in parts using natural
partitions such as parts, sections, chapters, paragraphs
or sentences.
Thus, we consider the corpus as a composite of $P$ parts.
For the $i$--th part of the corpus we can reckon up the total
number $N_i$ of tokens and the number $n_i(w)$
of occurrence of the word type $w$ inside this part.
Then, the fraction $f_i(w) = n_i(w)/N_i$ ($i = 1, \dots, P$)
is the relative frequency of occurrence of the word type $w$
in the part $i$.
Obviously, $\sum_{i=1}^P N_i = N$ is the total number of tokens
in the corpus and $\sum_{i=1}^P n_i(w) = n(w)$ is the number
of tokens corresponding to the word type $w$.
Therefore, it is possible to define a probability measure
over the partitions~\cite{MZ02} as
\begin{equation}
p_{i}(w) = \frac{f_{i}(w)}{\displaystyle\sum_{j=1}^P f_{j}(w)} \;.
\label{eq:prob_w_i}
\end{equation}
The quantity $p_i(w)$ results more complex than the conditional
probability $f_i(w)/(n(w)/N)$, of finding the word type $w$ in
the part $i$ given that it is present in the corpus.

Following Shannon's arguments, the information entropy
associated with the discrete distribution $p_i(w)$ is
\begin{equation}
S(w) = -\frac{1}{\ln(P)} \sum_{i=1}^P p_{i}(w) \ln(p_{i}(w)) \;.
\label{eq:entropy}
\end{equation}
The value $1/\ln(P)$ for the constant $K$ was selected
to take the maximum value of $S$ equal to one. Thus, $0< S(w) < 1$.
In this manner, when a type word is uniformly distributed
($p_i = 1/P$, for all $i$), Eq.~(\ref{eq:entropy}) yields $S = 1$.
Conversely, the other extreme case, $S = 0$, is when a word type
appears only in part $j$, thus we have $p_j=1$ and $p_i=0$
for $i \neq j$.
Therefore, words with frequent grammatical use like function words
(prepositions, adverbs, adjectives, conjunctions, and pronouns)
will have high values of entropy, meanwhile keywords will have
low values of entropy.
Empirical evidence~\cite{MZ02} shows a tendency of the entropy
to increase with $n$. It implies that, on average, the more
frequent word types are more uniformly used.

As we did with preceding indices, we need to calculate the
average of the entropy of a mock word type that appears $n$
times in a random corpus. From Eq.~(\ref{entropy_ran})
in Appendix~\ref{app:entropy}, we obtain
\begin{equation}
(1 - S)_{\mbox{ran}}  \approx
\frac{P-1}{2 \,n \,\ln P} \,,
\label{eq:entropy_ran}
\end{equation}
for $n >> 1$ and if all the parts of the random text have the same
number of tokens. Empirical evidence~\cite{MZ02} shows that the
agreement of Eq.~(\ref{eq:entropy_ran}) with random shuffling
of texts using natural partitions is very good, in spite of the
limitation of the last assumption.
From Eq.~(\ref{eq:entropy_ran}), we can see that the
dependence on the absolute frequency, $n$, is critical
for $(1 - S)_{\mbox{ran}}$ and it could not be ignored
even if the text is as large as a corpus.

\begin{figure}
\begin{center}
\includegraphics[clip,width=0.45\textwidth]{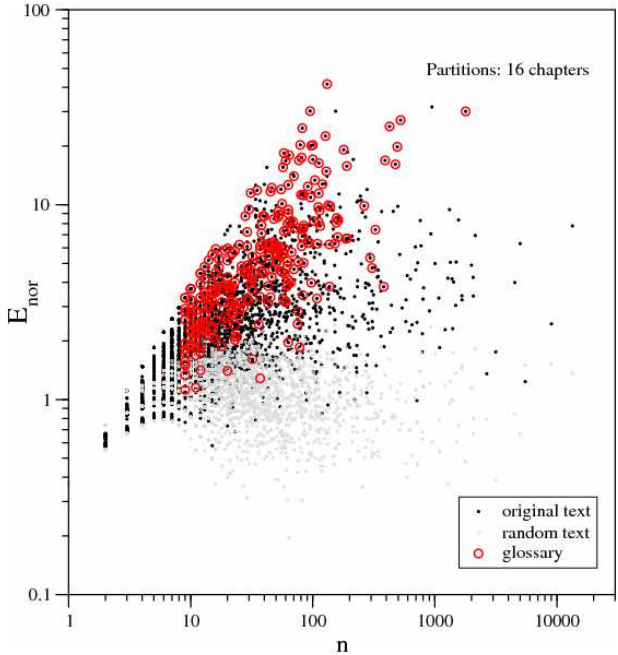}
\end{center}
\caption
{\small $E_{\mbox{nor}}$ vs.\ number of occurrence
($n$) for each word in Darwin's corpus.
We have also plotted superposed the random version of the
corpus (grey) and we have marked by open circles the words
corresponding to our prepared glossary (red online).}
\label{fig:entropy}
\end{figure}

Montemurro and Zanette~\cite{MZ02} proposed Eqs.~(\ref{eq:prob_w_i})
and (\ref{eq:entropy}) to study the distribution of words according
to their linguistic role. For this task, they found that the suitable
coordinates whereby words can be categorized are $n \,(1-S)$ and $n$.
In the same way, we will use these ideas for detecting relevance
of words. We cannot use directly the entropy as index because all
tokens with only one occurrence have zero entropy.
Thus, we define a normalised index freed from the
dependence on absolute frequency ($n$) in random texts by
\begin{equation}
E_{\mbox{nor}}(w) = n(w) \,(1 - S(w))_{\mbox{nor}} =
n(w) \frac{2 \,\ln P}{P-1} \,(1 - S(w))
\,.
\label{eq:Snor}
\end{equation}
Figure~\ref{fig:entropy} shows the values of $E_{\mbox{nor}}$
for all word types of Darwin's corpus versus its number
of occurrence, $n$, {\em on a double logarithmic scale}.
The individual deviations from the bulk trend for each value
of $n$ are related to the particular usage nuances of words.
To stress these deviations, we have used the 16 chapters of
the corpus as natural partitions for our entropic analysis
({\em i.e.} $P=16$).
In this way, we obtain a remarkable scattering of higher values
of $E_{\mbox{nor}}$ in the full range of number of occurrences.
A same entropic analysis using the $842$ paragraphs of Darwin's
corpus as partitions ({\em i.e.} $P=842$) generates a similar graph
that stresses the bulk trend, but the fluctuations are completely
smoothed.
Using the chapters as partitions ($P=16$) in Fig.~\ref{fig:entropy},
the ``cloud of points'' corresponding to the random version of
the corpus is distributed around the unitary value and the corpus
appears clearly more separated from the random text than with
previous indices.
Additionally, the words corresponding to the glossary have
systematically high values of the index $E_{\mbox{nor}}$.

To reinforce our graphical findings, in the following section
we perform a quantitative comparison among the indices
$\sigma_{\mbox{nor}}$, $\kappa_{\mbox{nor}}$, $\Gamma_{\mbox{nor}}$,
and $E_{\mbox{nor}}$ based on the power of each index
for discriminating the glossary from the bulk of words.

\section{Glossary as benchmark}
\label{sec:benchmark}

\begin{figure}
\begin{center}
\includegraphics[clip,width=0.45\textwidth]{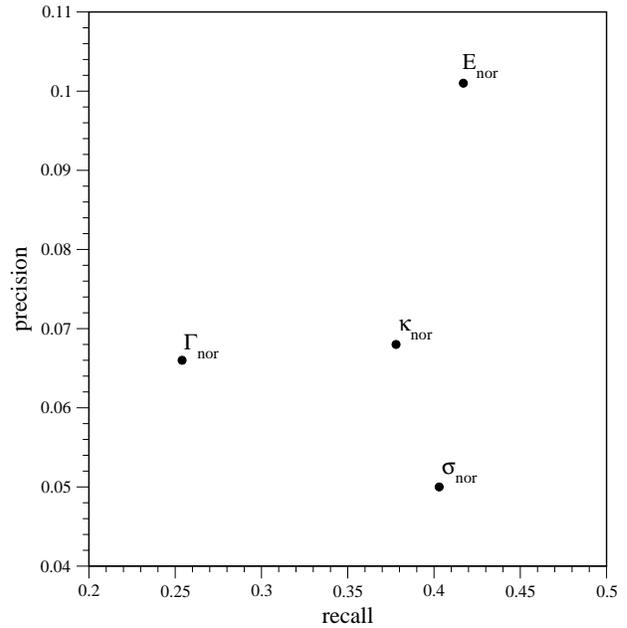}
\end{center}
\caption
{\small Comparison of information retrieval performance
of the indices (see Table~\ref{Table:Benchmarks}).
}
\label{fig:performance}
\end{figure}

Evaluation in information retrieval makes frequent use of
the notions of {\em recall} and {\em precision}~\cite{MS99,SM83}.
Recall is defined as the proportion of the target items that
a system recover. Precision is defined as a measure of the
proportion of selected items that are targets.
Remembering that our prepared glossary has $283$ word types,
we denote by $NG$ the number of the glossary's word types among
the first top $283$ ranked word types of the corpus.
For our purposes, we define recall of an index of relevance
as the fraction $NG/283$.
Thus, recall for the index $E_{\mbox{nor}}$ results $41 \%$.
On the other hand, precision can be built looking for the last
word type of our prepared glossary in the global ranking of each
index. For our convenience, we denote by $LP$ the position in
the ranking of the last word type of the glossary, and we define
precision of a keyword extractor as the fraction $283/LP$.
Thus, for example, the last entry of the glossary according to
the index $E_{\mbox{nor}}$ is {\small FLOWERING} and is ranked
in the position $2,790$. Remembering that the corpus has $8,294$
word types, we obtain that the complete prepared glossary is
allocated by $E_{\mbox{nor}}$ in the first third part of the
ranked lexicon and the precision of the index results $10 \%$.
Recall and precision are useful benchmarks for measuring the index's
performance. In particular, recall and precision of each index
analysed in this work are given in Table~\ref{Table:Benchmarks}.
We want to stress that the values of recall and precision of
the indices $\sigma$ and $\Gamma$ are exactly the same as those
obtained for $\sigma_{\mbox{nor}}$ and $\Gamma_{\mbox{nor}}$,
respectively. This fact is due to the normalisation factors
given by Eqs.~(\ref{eq:sigmaran}) and~(\ref{eq:Gammaran}), which
are almost constant for a corpus. Therefore, the pair of indices
$\sigma$ and $\sigma_{\mbox{nor}}$ (or $\Gamma$ and
$\Gamma_{\mbox{nor}}$) yield identical rankings of keywords.
\begin{table}
\caption
{\small Recall and precision of each index.
$NG$ is the number of glossary's word types among the first $283$
entries of each ranking.
$LP$ is the last position in each ranking in which appears
a word type of the glossary.
Thus, recall $= NG/283$ and precision = $283/LP$.}
\begin{center}
\vspace{5pt}
{\small
\begin{tabular}{lrcccrcl}
\hline
Index & $NG$ & recall & $LP$ & precision & last word \\
\hline
$E_{\mbox{nor}}$       & 118 & 0.417 & $2,790$ & 0.101 & FLOWERING \\
$\sigma_{\mbox{nor}}$  & 114 & 0.403 & $5,689$ & 0.050 & SCARCELY  \\
$\kappa_{\mbox{nor}}$  & 107 & 0.378 & $4,181$ & 0.068 & INDIAN    \\
$\Gamma_{\mbox{nor}}$  &  72 & 0.254 & $4,312$ & 0.066 & OSTRICH   \\
\hline
\end{tabular}
}
\end{center}
\label{Table:Benchmarks}
\end{table}
In order to compare the performance of all indices,
in Fig.~\ref{fig:performance} we have drawn a precision--recall
plot where we can see the significant improvement performed by
the index $E_{\mbox{nor}}$, both in recall and precision.
Also, in Fiq.~(\ref{fig:performance}) we see that
$\kappa_{\mbox{nor}}$ has a recall slightly worse than
$\sigma_{\mbox{nor}}$ and precision as good as
$\Gamma_{\mbox{nor}}$.
Thus, we find that the skewness of the distribution of occurrences
of a word type has a significant part of information about
the relevance of the word in the text.

In Table~\ref{Table:Rankins} we show the first top 50 word types
of the prepared glossary ranked by the index $E_{\mbox{nor}}$.
We also show the rank position of each word type by the others
indices.
\begin{table*}
\caption
{\small First top 50 word types of the prepared glossary ranked
by the index $E_{\mbox{nor}}$.
The numerical values correspond to the positions in the ranking of
each word type, not to the actual values of the indices.}

\begin{center}
\vspace{5pt}
{\small
\begin{tabular}{lrrrclrrr}
\hline
Word type    & $E_{\mbox{nor}}$ & $\sigma_{\mbox{nor}}$ & $\Gamma_{\mbox{nor}}$ &
\rule{0.3cm}{0cm}&
Word type    & $E_{\mbox{nor}}$ & $\sigma_{\mbox{nor}}$ & $\Gamma_{\mbox{nor}}$ \\
\hline
HYBRIDS      &    1    &   2   &  13     & &   SEA          &    33   &  65   &  309  \\
STERILITY    &    3    &   1   &   7     & &   SEEDS        &    35   &  64   &  279  \\
SPECIES      &    5    &  447  & 1312    & &   FERTILE      &    37   &  54   &  135  \\
FORMS        &    6    &  185  &  667    & &   ORGAN        &    39   &  14   &  218  \\
VARIETIES    &    7    &  39   &  384    & &   MOUNTAINS    &    40   &  120  &  94   \\
INSTINCTS    &    8    &   3   &  19     & &   GLACIAL      &    41   &  51   &  113  \\
BREEDS       &    9    &  38   &  142    & &   GARTNER      &    43   &  36   &  20   \\
FERTILITY    &    10   &   8   &  33     & &   HYBRID       &    44   &  46   &  59   \\
FORMATIONS   &    11   &  20   &  78     & &   CUCKOO       &    47   &  13   &   3   \\
CROSSED      &    12   &   9   &  82     & &   LAND         &    48   &  106  &  613  \\
SELECTION    &    13   &  212  &  858    & &   EGGS         &    50   &  109  &  215  \\
ORGANS       &    14   &  61   &  433    & &   STRUGGLE     &    51   &  829  &  571  \\
NEST         &    16   &  22   &  18     & &   BREED        &    52   &  332  &  367  \\
INSTINCT     &    17   &   5   &  32     & &   GEOLOGICAL   &    54   &  129  &  456  \\
RUDIMENTARY  &    18   &  25   &  130    & &   CROSS        &    62   &  125  &  205  \\
FORMATION    &    19   &  144  &  341    & &   HABITS       &    63   &  278  & 1260  \\
BEES         &    21   &   6   &  29     & &   STRUCTURE    &    65   &  105  & 1451  \\
PLANTS       &    22   &  113  &  776    & &   INHABITANTS  &    67   &  95   &  556  \\
CELLS        &    23   &  18   &  50     & &   FLOWERS      &    68   &  35   &  250  \\
POLLEN       &    24   &  12   &  74     & &   ANTS         &    75   &  41   &  35   \\
NATURAL      &    25   &  460  & 1288    & &   RACES        &    78   &  566  &  542  \\
GROUPS       &    26   &  79   &  393    & &   OFFSPRING    &    81   &  400  &  884  \\
CROSSES      &    27   &  60   &  81     & &   SEXUAL       &    85   &  89   &  285  \\
WATER        &    29   &  75   &  400    & &   VARIABLE     &    87   &  138  &  467  \\
STERILE      &    31   &  19   &  109    & &   WILD         &    89   &  235  &  269  \\
\hline
\end{tabular}
}
\end{center}
\label{Table:Rankins}
\end{table*}
A false positive is when the system identifies a keyword that
really is not one. In Table~\ref{Table:False positives} we
show the first top 40 ranked (by $E_{\mbox{nor}}$) word types
not included in our prepared glossary. We can immediately
see that several terms are not necessarily false positives.
We have marked with an asterisk (*) in the table those word
types that were not previously selected in the prepared glossary,
but that appeared in the main entries of the original glossary
of Darwin's book. Indeed, several more word types like these
could have been included in our prepared glossary, too.
Moreover, we could say that the word type {\small I} is relevant for
a text that uses the first--person narrative, like Darwin's book.
{\small ISLAND} and {\small SLAVES} were not used neither
in the book's glossary nor in its index;
however $E_{\mbox{nor}}$ ranks it adequately as a keyword.
The word type {\small F} is also meaningful to the text.
It appear in the proper nouns ``Mr. F. Smith'' and
``Dr. F. Muller'', and in the collocations ``F. sanguinea'',
``F. rufescens'', ``F. fusca'', ``F. flava'', and ``F. rufescens''
which denote species.
\begin{table*}
\caption
{\small First 40 false positives word types ranked by
the index $E_{\mbox{nor}}$ and its numbers of occurrences $n$.
The numerical values in the $E_{\mbox{nor}}$ column correspond
to the positions in the ranking of each word types, not to the
actual values of the index.}

\begin{center}
\vspace{5pt}
{\small
\begin{tabular}{lrrrclrrr}
\hline
Word type    & $E_{\mbox{nor}}$ & $n$ & &
\rule{0.3cm}{0cm}&
Word type    & $E_{\mbox{nor}}$ & $n$ & \\
\hline
I             &     2   &    947   &    & &   NORTHERN      &   60   &     41  &    \\
ISLANDS       &     4   &    154   & *  & &   DESCENT       &   61   &     80  & *  \\
CHARACTERS    &    15   &    192   & *  & &   FRESH         &   64   &     50  & *  \\
GENERA        &    20   &    215   & *  & &   ITS           &   66   &    497  &    \\
WAX           &    28   &     42   &    & &   DIFFERENCES   &   69   &    168  &    \\
ISLAND        &    30   &     69   &    & &   CELL          &   70   &     30  &    \\
DOMESTIC      &    32   &    131   & *  & &   EXTINCT       &   71   &    116  & *  \\
YOUNG         &    34   &    127   &    & &   EUROPE        &   72   &     81  & *  \\
TEMPERATE     &    36   &     40   &    & &   FERTILISED    &   73   &     34  &    \\
SLAVES        &    38   &     34   &    & &   DIAGRAM       &   74   &     40  &    \\
NEW           &    42   &    278   &    & &   SHALL         &   76   &    105  &    \\
MY            &    45   &     99   &    & &   WE            &   77   &   1320  &    \\
INCREASE      &    46   &     82   &    & &   DEVELOPED     &   79   &    146  & *  \\
INTERMEDIATE  &    49   &    164   &    & &   BEDS          &   80   &     35  &    \\
PERIOD        &    53   &    245   & *  & &   ADULT         &   82   &     46  &    \\
MIVART        &    55   &     34   & *  & &   TWO           &   83   &    456  &    \\
THROUGH       &    56   &    249   &    & &   BETWEEN       &   84   &    367  &    \\
HE            &    57   &    236   &    & &   NUMBER        &   86   &    255  &    \\
F             &    58   &     37   &    & &   OCEANIC       &   88   &     42  & *  \\
PARTS         &    59   &    230   & *  & &   THEORY        &   90   &    131  &    \\
\hline
\end{tabular}
}
\end{center}
\label{Table:False positives}
\end{table*}
The observations in the last paragraphs induce us to consider that
the performance of the index $E_{\mbox{nor}}$ is better than what
can be inferred from Table~\ref{Table:Benchmarks}.

Moreover, the index $E_{\mbox{nor}}$ requires less computational
efforts that the others. Knowing the number of occurrences of
a word type, the implementation of the algorithm for the variance
or the skewness requires of one accumulator plus a counter for
reckoning the number of tokens between nearest neighbour
occurrences of the word type.
While, for the entropic index, we only need one counter
(of number of occurrences) for each partition per word type.
On the other hand, the algorithm for $\Gamma$ requires three
accumulators and for each occurrence of a word type we need
to determine if it corresponds to a cluster point.

\section{Concluding remarks}
\label{sec:fin}

In summary, in this work we addressed the issue of statistical
distribution of words in texts. Particularly, we have concentrated
on the statistical methods for detecting keywords in literacy text.
We reviewed two indices ($\sigma$ and $\Gamma$) previously
proposed~\cite{OCB+02,ZS03} for measuring relevance and we
improved them by considering their values in random texts.
Additionally, we introduced $\kappa_{\mbox{nor}}$ based on
the skewness of the distribution of occurrences of a word and
we proposed another index for keyword detection based on the
information entropy. Our proposals are very easy to implement
numerically and have performances as detectors as good as
or better than the other indices.
The ideas of this work can be applied to any natural language
with words clearly identified, without requiring any
previous knowledge about semantics or syntax.

\section*{Acknowledgements}

Contributions to Appendix~\ref{app:entropy}
by Marcelo Montemurro are gratefully acknowledged.
This work was partially supported by grant from
``Se\-cre\-ta\-r\'\i a de Cien\-cia y Tec\-no\-lo\-g\'\i a
de la Uni\-ver\-si\-dad Na\-cio\-nal de C\'or\-doba''
(Code: 05/B370).

\appendix

\section{The Geometrical distribution}
\label{app:pascal}

In this Appendix we briefly review the basic results of the
geometrical distribution, scattered in the literature, that
are useful for this work.
First, we consider an experiment with only two possible outcomes
for each trial (binomial experiment).
Repeated independent trials of the binomial experiment are called
Bernoulli trials if their probabilities remain constant throughout
the trials.
We denote by $p$ the probability of the ``successful'' outcome.
Now, we are interested in the probability of success on the
$j$--th trial after a given success. Given that the trials are
independent, we immediately obtain the geometrical distribution
\begin{equation}
P(j) = (1-p)^{j-1} \,p \,, \qquad \mbox{for }j \geq 1 \;.
\label{geometrica}
\end{equation}
%

\subsection{Moments and cumulants}

The characteristic function of a stochastic variable $X$ is
defined by
$G(k) = \left< e^{k X} \right>
      = \sum_{j \geq 1} P(j) \,\exp(k j)$.
Thus, for the geometrical distribution we obtain
\begin{equation}
G(k) = \displaystyle\frac{p \,e^k}{1 - (1-p) \,e^k} \;.
\label{G}
\end{equation}
This function is also the moment generating function
\begin{equation}
\left< X^n \right> =
\left. \frac{d^nG}{dk^n} \right|_{k=0} \,.
\label{moments}
\end{equation}
Therefore, the first three cumulants of the geometrical
distribution are given by
\begin{equation}
\begin {array}{l}
\mu _1 = \left< X \right> =
\displaystyle\frac{1}{p} \;,\\
\mu _2 = \left< X^2 \right> - \left< X \right>^2 =
\displaystyle \frac{1-p}{p^2} \;,\\
\mu _3 = \left< X^3 \right>
- 3 \,\left< X^2 \right>  \,\left< X \right>
+ 2 \,\left< X \right>^3 =
\displaystyle\frac{(2-p) \,(1-p)}{p^3} \;.
\end{array}
\label{cumulants}
\end{equation}
%

\subsection{Addition of two geometrical variables}

If $X_1$ e $X_2$ are geometrical distributed independent random
variables, the distribution of the addition $Y = X_1 + X_2$ is
\begin{equation}
P_Y(j) = \sum_{m_1+m_2=j} P(m_1,m_2) \;,
\qquad \mbox{for } j = 2,3, \dots \;,
\end{equation}
where the joint probability distribution of the variables
$X_1$ e $X_2$, $P(m_1,m_2)$, is given by
\begin{equation}
P(m_1,m_2) = p^2 \,(1-p)^{m_1+m_2-2} \,,
\mbox{for } m_1 \geq 1,\mbox{and } m_2 \geq 1 \;.
\end{equation}
In this manner,
\begin{equation}
P_Y(j) = \sum_{m=1}^{j-1} P(m,j-m) =
\sum_{m=1}^{j-1} \,p^2 \,(1-p)^{j-2} \;.
\end{equation}
Therefore
\begin{equation}
P_Y(j) = (j-1) \,p^2 \, (1-p)^{j-2} \;, \;\;
\mbox{for } j = 2,3, \dots \;.
\label{sumageo}
\end{equation}

Now, we are interested in the average of the random variable
(recall Eq.~(\ref{eq:cluster}))
\begin{equation}
\gamma = \left\{
\begin{array}{lcl}
1 - \displaystyle\frac{Y}{2 \,\mu} \,, && Y < 2 \mu \\
0 \,, && Y \geq 2 \mu
\end{array}
\right. \;,
\label{gamma}
\end{equation}
where $Y$ is the addition of two independent geometrical distributed
random variables with mean $\mu = 1/p$. By definition we have that
\begin{equation}
\left< \gamma \right> = \sum_{j=2}^h
\left( 1 - \displaystyle\frac{j}{2 \,\mu} \right) \,P_Y(n) \;,
\label{avg_gamma}
\end{equation}
where $P_Y(n)$ is given by Eq.~(\ref{sumageo})
and  $h = \mbox{Int}[2 \mu]$.
Defining $q = 1-p$ and using the identity
\begin{equation}
\sum_{n=1}^N q^n = \frac{q - q^{N+1}}{1 - q}
\end{equation}
we immediately obtain
\renewcommand{\arraystretch}{1.5}
\begin{equation}
\displaystyle
\sum_{n=j}^h P_Y(j)
= \displaystyle
p^2 \frac{d}{dq} \sum_{k=2}^h q^{k-1} \\
= 1 - h \,q^{h-1} + (h-1) \,q^h \;,
\end{equation}
and
\begin{equation}
\begin{array}{l}
\displaystyle
p \,\sum_{j=2}^h  j \,P_Y(j)
=
\displaystyle
p^3 \frac{d^2}{dq^2} \sum_{k=2}^h q^{k}
= 2 - h \,(h+1) \,q^{h-1} \\+ 2 \,(h+1) \,(h-1) \,q^h
- h \,(h-1) \,q^{h+1} \;.
\end{array}
\end{equation}
Therefore
\begin{equation}
\left< \gamma \right> =
\displaystyle\frac{1}{2} \,h \,(h-1) \,q^h \,(q + q^{-1} - 2) \;.
\label{gammaran}
\end{equation}

The Poisson distribution can be obtained from the geometrical
distribution in the limit $p \rightarrow 0$.
Expanding $q^z$ into a Taylor series up to fourth order
we obtain
\begin{equation}
q^{h+1} + q^{h-1} - 2 \,q^h \approx
p^2 + (1-h) \,p^3 + \frac{1}{2} \,(2-3h+h^2) \,p^4 \;.
\end{equation}
Given that for $p \rightarrow 0$ we have $h >> 1$,
the last equation can be recast as
\begin{equation}
\begin{array}{l}
q^{h+1} + q^{h-1} - 2 \,q^h \approx
p^2 \,\left( 1 - h \,p + \frac{1}{2} (hp)^2  \right)
\\ \approx p^2 \exp{(-hp)} \;.
\end{array}
\end{equation}
Finally, using that $hp \approx 2$, we obtain that the
average of the random variable $\gamma$ for a Poisson
distribution~\cite{ZS03} is
\begin{equation}
\left< \gamma \right> = 2 \,e^{-2} \;.
\label{Poisson}
\end{equation}
%

\section{Entropy of a random text}
\label{app:entropy}

Here, we derive the entropy of a random text in a more detailed
way that is described in Ref.~\cite{MZ02}.

We consider a corpus of $N$ tokens as a composite of $P$ parts,
with $N_i$ tokens in the $i$--th part ($i = 1, 2, \dots, P$).
In a random corpus, the probability that a word type $w$ appears
in the part $j$ is $N_j/N$. Thus, the probability that $w$ appears
$n_1$ times in part 1, $n_2$ times in part 2, and so on, is the
multinomial distribution
\begin{equation}
p_{w}(n_{1}, n_{2}, \ldots, n_{P}) = n! \,
\prod_{j=1}^P \frac{1}{n_{j}!}
\left( \frac{N_{j}}{N} \right)^{n_{j}} \,,
\label{prob_n_i}
\end{equation}
where $n = \sum_{j=1}^P n_j$ is the absolute frequency
(number of tokens) of the word type $w$.

For reasons of simplicity, in this Appendix we consider the
particular case in which all the parts have exactly the same
number of tokens, {\em i.e.} $N_i = N/P$. Hence, the probability
measure defined by Eq.~(\ref{eq:prob_w_i}) can be simply written as
$p_i = n_i/n$ and the information entropy defined by
Eq.~(\ref{eq:entropy}) results
\begin{equation}
S = - \frac{1}{\ln P}
\sum_{i=1}^P \frac{n_i}{n} \,\ln \left( \frac{n_i}{n} \right) \,.
\label{entropy}
\end{equation}

Now, we are interested in the average value of the entropy over
the distribution given by Eq.~(\ref{prob_n_i}). We only need to
compute the average of each term of Eq.~(\ref{entropy}) using
the marginal distributions, $p_w(n_i)$, obtained from
Eq.~(\ref{prob_n_i}). All marginal distributions result
binomials with mean $n/P$ and variance $n/P(1-1/P)$.
Thus, we obtain for the average entropy
\begin{equation}
\left< S \right> =
-\frac{P}{\ln P} \sum_{m=0}^n \frac{m}{n}
\ln \left( \frac{m}{n} \right)
\left( {n}\atop{m} \right)
\frac{1}{P^{m}} \left( 1-\frac{1}{P} \right)^{n-m} \,.
\label{<entropy>}
\end{equation}

For highly frequent word types, $n >> 1$, we can approximate
the binomial distribution by a Gaussian probability function
($G(x;\mu,\sigma)$) with mean $\mu = 1/P$ and variance
$\sigma^2 = (1/n)(P-1)/P^2$.
Thus, Eq.~(\ref{<entropy>}) can be recast as
\begin{equation}
\left< S \right> \approx
-\frac{P}{\ln P} \int_0^1 x \ln x \,G(x;\mu,\sigma) dx \,.
\label{approx<entropy>}
\end{equation}
In the limit $n>>1$, $\sigma \rightarrow 0$ and the Gaussian
probability function concentrates around its mean value $\mu$.
Using the expansion of the function $x \ln x$ around $\mu$,
\begin{equation}
x \ln x \approx \mu \ln \mu + (1 + \ln \mu) (x-\mu)
+\frac{1}{2} \frac{1}{\mu} (x-\mu)^2 \,,
\label{expansion}
\end{equation}
in Eq.~(\ref{approx<entropy>}) and remembering that
$$\int_{-\infty}^{\infty} (x-\mu)^2 \,G(x;\mu,\sigma) dx
= \sigma^2 \,,$$
we finally obtain for a random text~\cite{MZ02} that
\begin{equation}
\left< S \right> \approx
1 - \frac{P-1}{2 \,n \,\ln P} \,.
\label{entropy_ran}
\end{equation}
%

%

\end{document}